\documentclass[sigconf,nonacm]{mechscampreprint}

\AtBeginDocument{%
  }

\setcopyright{none}
\renewcommand{\footnotetextcopyrightpermission}[1]{}

\makeatletter
\renewcommand{\@authorfont}{\normalfont\fontsize{12}{14}\selectfont}
\renewcommand{\@affiliationfont}{\normalfont\fontsize{10}{12}\selectfont}
\makeatother

\usepackage{array}
\usepackage{tabularx}
\usepackage{colortbl}
\usepackage{xspace}
\definecolor{tablegroupgray}{gray}{0.90}
\newcommand{\tablegroup}[2]{%
  \rowcolor{tablegroupgray}[0pt][0pt]%
  \multicolumn{#1}{@{}c@{}}{\strut\textbf{#2}}\\}

\begin{document}

\title{Beyond Risk Cues: A Mechanism-Level Benchmark and Verifiable Debate for Scam Video Detection on the Web}

\author{Yifan Wang}
\affiliation{%
  \institution{People's Public Security\\University of China}
  \city{Beijing}
  \country{China}}
\email{2024111026@stu.ppsuc.edu.cn}

\author{Junyu Lu}
\affiliation{%
  \institution{Dalian University of Technology}
  \city{Dalian}
  \country{China}}
\email{dutljy@mail.dlut.edu.cn}

\author{Shun Zhang}
\affiliation{%
  \institution{People's Public Security\\University of China}
  \city{Beijing}
  \country{China}}
\email{shunzhang@ppsuc.edu.cn}

\author{Chaozhuo Li}
\affiliation{%
  \institution{Microsoft}
  \country{China}}
\email{lichaozhuo@bupt.edu.cn}

\author{Fanliang Bu}
\affiliation{%
  \institution{People's Public Security\\University of China}
  \city{Beijing}
  \country{China}}
\email{bufanliang@sina.com}

\author{Rong Yin}
\affiliation{%
  \institution{Beihang University}
  \city{Beijing}
  \country{China}}
\email{yinrong@buaa.edu.cn}

\author{Lingbin Bu}
\affiliation{%
  \institution{Beijing Police College}
  \city{Beijing}
  \country{China}}
\email{bulingbin@bjpc.edu.cn}

\author{Changrong Min}
\affiliation{%
  \institution{Criminal Investigation Police\\University of China}
  \city{Shenyang}
  \country{China}}
\email{mcr19940816@gmail.com}

\author{Peng Chen}
\affiliation{%
  \institution{People's Public Security\\University of China}
  \city{Beijing}
  \country{China}}
\email{chenpeng@ppsuc.edu.cn}

\renewcommand{\shortauthors}{Wang et al.}

\begin{abstract}
Online scam videos pose significant threats to individuals and society. Existing studies primarily focus on binary classification, offering limited insights into how scams work and the reasoning behind detection decisions. Moreover, they often overlook the challenge of distinguishing scam videos from superficially similar non-scam content, such as advertisements, and the robustness of detection models against misleading arguments. These limitations hinder reliable scam video detection.
To address these limitations, we introduce \textbf{MechScam}, a benchmark comprising 6,614 videos covering diverse scam mechanisms and non-scam categories. MechScam provides explanations supported by video evidence, annotations of potential scam elements and their context, and misleading arguments constructed from genuine evidence. Recognizing that genuine evidence can still be used to justify misleading conclusions, we further propose \textbf{ScamCourt}, a debate framework in which trained prosecution and defense debaters present opposing arguments, while a trained judge makes decisions based on a transcript with verified citations. Extensive experiments show that state-of-the-art MLLMs struggle to distinguish scam videos from similar non-scam content. ScamCourt outperforms larger MLLMs and same-backbone fine-tuned baselines, reducing false positives and showing stronger resistance to misleading arguments. Together, MechScam and ScamCourt contribute to more reliable and interpretable scam video detection. The dataset and code are available at https://github.com/SuperYFan/MechScam.
\end{abstract}



\makeatletter
\let\preprint@keywords\@keywords
\let\@keywords\@empty
\makeatother
\maketitle
\hypersetup{pdfauthor={Yifan Wang, Junyu Lu, Shun Zhang, Chaozhuo Li, Fanliang Bu, Rong Yin, Lingbin Bu, Changrong Min, Peng Chen}}

\makeatletter
\let\@keywords\preprint@keywords
\hypersetup{pdfkeywords={\@keywords}}
\par\vfill
\noindent\begin{minipage}{\columnwidth}
    \normalfont\normalsize
    {\bfseries CCS Concepts\par}
    \smallskip
    \noindent\@concepts\par
    \medskip
    {\bfseries Keywords\par}
    \smallskip
    \noindent\@keywords\par
\end{minipage}
\par\newpage
\makeatother

\begin{figure}[H]
    \centering
    \includegraphics[width=\columnwidth]{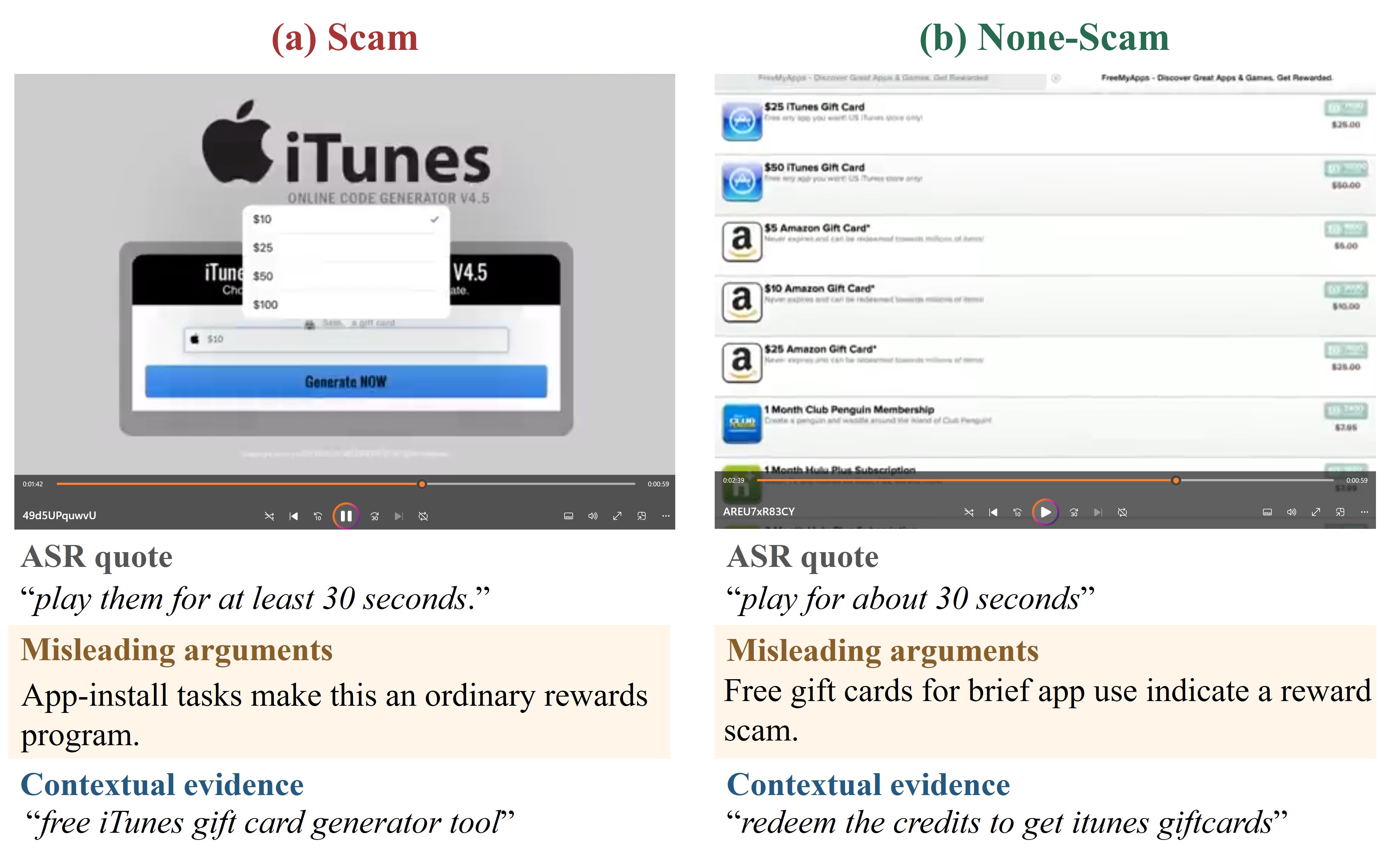}
    \caption{Scam and non-scam examples with similar cues.}
    \label{fig:motivation}
\end{figure}

\section{Introduction}

Video-sharing platforms are an important channel for information consumption and a venue for scam promotion~\cite{boumasims2025kids}.
Scam videos combine visual demonstrations, spoken claims, and textual instructions to make reward, investment, and recruitment schemes appear credible.
Return figures, referral links, and participation requests also occur in ordinary promotions, tutorials, market commentary, and anti-scam investigations.
As illustrated in Figure~\ref{fig:motivation}, both the scam video (a) and non-scam video (b) link free iTunes gift cards to roughly 30 seconds of app use.
In (a), a purported gift-card generator requires app installation to unlock codes.
In (b), app offers earn credits redeemable through a reward catalog.
These shared reward and action cues make the videos easy to confuse; distinguishing them requires identifying how each reward is obtained.

Existing datasets support multimodal scam detection and, in some cases, policy-based explanations, with limited mechanism-level annotations spanning scam content and benign look-alikes~\cite{vidscamnet,youtubescam}.
Evaluation centered on binary labels can overlook whether models identify how a scheme operates or rely on familiar topics and risk cues, reflecting the broader challenge of shortcut learning~\cite{geirhos2020shortcut}.
A model may correctly flag an investment promotion as a scam while confusing platform lending with automated trading, leaving the underlying mechanism unexplained.
Reward claims and financial participation occur in both deceptive solicitations and ordinary promotions or tutorials, making context essential for assessing deception.
Fine-grained categories and contextually similar benign examples are therefore needed to assess both mechanism attribution and the boundaries of scam judgments.

Evidence-linked explanations introduce a further challenge.
Selectively emphasizing app use can portray (a) as an ordinary rewards program or (b) as a reward scam.
Both arguments can cite genuine statements while overlooking the mechanism that gives those statements meaning.
Assessing whether a conclusion follows from the evidence requires examining how the cited content is interpreted.
Recent approaches use multi-agent debate~\cite{mvdebate2025,d2d2025} or adversarial multi-stage reasoning~\cite{mars2026} to support content judgments. Their judges must resolve competing interpretations of the same evidence.
Evaluation should also examine whether models maintain correct judgments when exposed to misleading arguments grounded in genuine sources.

To address this, we introduce MechScam, a mechanism-level benchmark for evaluating scam mechanism attribution and discrimination of benign look-alikes.
It combines fine-grained labels with source-linked rationales, contextualized risk elements, and misleading arguments constructed from genuine video evidence.
These annotations characterize the relationship between promised benefits and requested actions while preserving the context that distinguishes promotion from demonstration, discussion, or criticism.
Building on MechScam, we propose ScamCourt, a verifiable-citation debate framework with trained prosecution, defense, and judging roles.
The prosecution and defense develop opposing arguments from the same multimodal sources, making disagreements about evidence interpretation explicit.
Text quotations and frame references are checked before judgment, establishing source provenance while leaving the judge to assess whether the evidence supports the proposed label and category.
We train the debaters on teacher-generated arguments for opposing labels and train the judge on debates generated through cross-fitting.

Our main contributions are summarized as follows:
\begin{itemize}
    \item We introduce MechScam, a benchmark of 6,614 YouTube videos across six scam mechanisms and six benign categories, annotated with evidence-grounded rationales and contextualized risk elements for evaluating mechanism attribution and benign look-alikes.

    \item We propose ScamCourt, a verifiable-citation debate framework with trained prosecution and defense roles and a judge trained on cross-fitted debates to evaluate competing interpretations of multimodal evidence.

    \item Our experiments reveal systematic errors on benign look-alikes and susceptibility to misleading arguments grounded in genuine evidence, while demonstrating the effectiveness of ScamCourt.
\end{itemize}

\section{Related Work}
\label{sec:related}
\subsection{Scam and Harmful Video Benchmarks}
Early YouTube scam studies examine giveaway, monetary, and arbitrage-bot scams, primarily through textual and statistical signals~\cite{boumasims2021youtube,tripathi2022monetizing,li2024arbitrage}. Measurement studies further investigate cryptocurrency giveaway operations, victim conversion, and user-reported scams~\cite{li2023giveaway,liu2024giveaway,extra1}. VidScam provides full video content and policy-grounded annotations, with VidScamNet producing multimodal predictions and policy-aligned explanations~\cite{vidscamnet}, while YouTube\_Scam supplies cryptocurrency-related cases with metadata and URL annotations~\cite{youtubescam}.

Harmful-video benchmarks cover multimodal and implicit hate detection~\cite{das2023hatemm,extra2,rehman2025implihatevid}. Closest to our annotation design, IARE augments hate-video datasets with captions, potentially harmful elements, and contextual rationales~\cite{lu2026iare}. MetaHarm, SafeWatch-Bench, and HarmVideoBench extend coverage to broader harms, fine-grained safety, and evidence-based understanding~\cite{jo2025metaharm,chen2025safewatch,wu2026harmvideobench}. For misinformation, FakeSV and FakeTT support short-video detection~\cite{qi2023fakesv,bu2024fakingrecipe}, while MisVideoQA~\cite{misvideoqa} and WildFakeBench~\cite{zeng2026wildfake} extend evaluation to fine-grained categories and reasoning; related resources examine video intent~\cite{atreya2025intenthq}. MechScam focuses on scam mechanisms and benign look-alikes, relating promised benefits, requested actions, and claimed return sources to evidence in context. Its annotations support joint evaluation of mechanism attribution and false positives on benign content sharing similar risk cues.

\subsection{LLM-based Detection and Debate}
\paragraph{LLMs for scam and harmful-content detection.}
General-purpose MLLMs~\cite{qwen3vl}, LLM-based moderation~\cite{kumar2024moderation}, and comparisons with fine-tuned scam classifiers~\cite{extra3} explore content-based detection. For explainable video detection, VidScamNet uses policy-grounded supervision, IARE applies information augmentation and preference optimization, and SafeWatch learns policy-following judgments with explanations. Prompted multi-agent approaches further explore opposing hypotheses, complementary views, and evidence integration for harmful-content and misinformation detection.

\paragraph{Debate as a judging protocol.}
Debate was proposed to help weaker judges supervise stronger agents~\cite{irving2018debate}. Subsequent studies examine persuasive debaters~\cite{khan2024debating}, unreliable experts~\cite{michael2023debate}, self-play training~\cite{arnesen2024selfplay}, and multi-agent reasoning~\cite{du2024multiagent,liang2024encouraging}. \citet{khan2024debating} verify quotations against source passages, providing a precedent for our citation checks. ScamCourt combines trained prosecution and defense roles with a judge trained on cross-fitted debates, checking both text quotations and frame references. Our robustness evaluation examines how models interpret genuine video evidence when presented with misleading arguments. Sycophancy concerns agreement with a user's stated belief~\cite{sharma2024sycophancy}. Our setting tests susceptibility to selective evidence and misleading interpretation.

\section{MechScam Curation}
\label{sec:dataset}

\begin{figure*}[t]
\centering
\includegraphics[width=\textwidth]{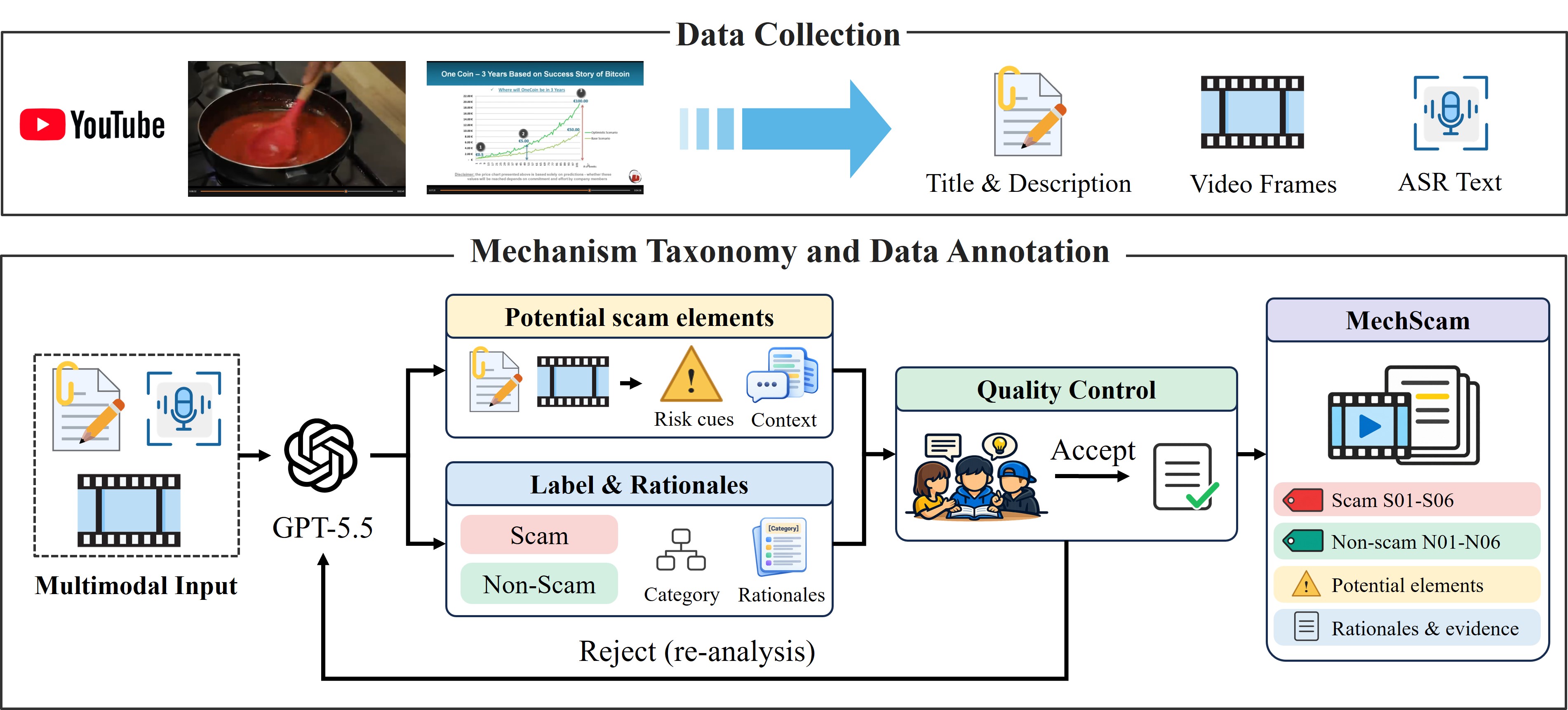}
\caption{Overview of our MechScam benchmark: Data Collection, Mechanism Taxonomy and Data Annotation.}
\label{fig:dataset}
\end{figure*}

\subsection{Data Collection}
We construct MechScam by collecting accessible videos from VidScam~\cite{vidscamnet} and YouTube\_Scam~\cite{youtubescam}, excluding those whose original videos are no longer available.
The resulting corpus contains 6,614 videos, with 3,079 from VidScam and 3,535 from YouTube\_Scam, all re-annotated under a unified framework.
Figure~\ref{fig:dataset} summarizes the collection and annotation pipeline.
Each video includes its title, uploader-provided description, speech transcript, and up to 64 uniformly sampled frames, yielding 418,442 frames in total.
We use Whisper large-v3 to transcribe English speech and translate non-English speech into English.

\subsection{Mechanism Taxonomy}
\label{sec:taxonomy}
We develop a taxonomy of six scam mechanisms and six non-scam content categories through manual analysis of the collected videos, as shown in Table~\ref{tab:taxonomy}.
The design draws on prior characterizations of cryptocurrency scams, giveaway schemes, trading-bot scams, and recruitment-based schemes~\cite{bartoletti2021scams,li2023giveaway,li2024arbitrage,kell2023forsage}, with category definitions adapted to the mechanisms observable in our video corpus.
We examine how promised benefits relate to requested actions and claimed sources of returns within the video's communicative context.
These relationships help distinguish deceptive solicitation from ordinary promotions, demonstrations, and discussions that share similar financial claims or reward-related cues.

Scam categories describe the mechanism connecting the promised benefit to the viewer's participation.
S01 covers purported reward or asset acquisition, while S06 covers commissions tied to recruitment and downline participation.
S02--S05 distinguish investment mechanisms by the advertised basis of returns: automated trading (S02), platform deposits (S03), purchased mining or node resources (S04), and token appreciation or holding-based dividends (S05).
Videos involving different assets, brands, or platforms can therefore share a category when they present the same mechanism.

Non-scam categories reflect the video's primary theme and purpose, distinguishing scam awareness and investigation, financial analysis, education and tutorials, products and services, rewards and promotions, and general content.
Each video receives one primary category reflecting its dominant mechanism or theme.
Appendix~\ref{app:category-examples} illustrates all twelve categories with real samples.

\begin{table*}[t]
\caption{The MechScam taxonomy: category names, descriptions, and video counts by primary category. S and N denote scam and non-scam categories, respectively.}
\label{tab:taxonomy}
\centering
\small
\begin{tabular*}{\textwidth}{@{\extracolsep{\fill}}l>{\raggedright\arraybackslash}p{0.24\textwidth}>{\raggedright\arraybackslash}p{0.57\textwidth}r@{}}
\toprule
ID & Category & Description & Videos \\
\midrule
S01 & Reward Claims & Claims free rewards, gift cards, or game assets in exchange for participation. & 214 \\
S02 & Trading Bots & Promises returns from trading bots or automated trading contracts that viewers authorize. & 363 \\
S03 & Platform Investments & Promises investment, lending, or staking returns on funds deposited with a platform. & 325 \\
S04 & Mining and Nodes & Promises returns from purchased mining capacity, mining packages, or node participation. & 448 \\
S05 & Token Investments & Promotes token purchases with promises of appreciation or holding-based dividends. & 327 \\
S06 & Recruitment Commissions & Promises commissions tied to recruiting others, downlines. & 507 \\
\midrule
N01 & Scam Awareness and Investigation & Warns about, investigates, tests, or exposes scams. & 294 \\
N02 & Financial News and Analysis & Reports financial events or analyzes markets and investment trends. & 1,428 \\
N03 & Education and Tutorials & Teaches concepts, technical skills, or operational procedures. & 1,351 \\
N04 & Products and Services & Introduces or sells ordinary products and services. & 363 \\
N05 & Rewards and Promotions & Presents ordinary giveaways, discounts, cashback, loyalty points. & 351 \\
N06 & General Content & Covers other news, lifestyle, entertainment, sports, or general-interest topics. & 643 \\
\midrule
\multicolumn{3}{@{}l}{Scam / Non-scam} & 2,184 / 4,430 \\
\bottomrule
\end{tabular*}
\end{table*}

\subsection{Data Annotation}
\label{sec:annotation}
We adopt a model-drafted, human-verified annotation workflow comprising potential scam element extraction, label and rationale annotation, as illustrated in Figure~\ref{fig:dataset}.
GPT-5.5 drafts the annotations, and every draft is verified by human annotators before finalization.

\paragraph{Potential scam elements.}
We separately examine textual sources and sampled video frames to identify potentially relevant claims, requests, and visual cues.
Each candidate records an observation, its supporting evidence, the surrounding context, and the limitations of what the supplied material establishes.
Candidates must capture a meaningful relationship, such as a promised benefit linked to a payment or recruitment requirement; isolated mentions of money, brands, or external links are insufficient.
Warnings, quotations, demonstrations, and other qualifying context are retained to distinguish the presence of a risk cue from its interpretation.

\paragraph{Labels, categories, and rationales.}
Annotation draws on video frames, titles, uploader-provided descriptions, and speech transcripts, with the taxonomy definitions and a fixed research guide derived from YouTube's spam and scam policy.
Neutral video captions provide auxiliary summaries of the original material.
Each video receives a binary label, a primary category, and annotations of its communicative role, such as promotion, warning, or discussion.
A contextual rationale explains the binary decision, while a fine-grained rationale explains the category assignment.
Both are grounded in quotations from identified text sources or references to supplied video frames.
Annotators distinguish a creator's own promotion from reported or investigated schemes and assess the evidence for deception within each video's communicative context.
All annotations undergo automatic consistency and citation checks, independent human review, and consensus-based correction, as detailed in Appendix~\ref{app:quality-control}.

\subsection{Misleading Argument Set}
\label{sec:adv-set}
We construct an misleading argument set to evaluate whether models can resist misleading interpretations of genuine video evidence.
For each video, GPT-5.5 uses the original multimodal content and reference annotations to generate three arguments supporting the opposite label.
These arguments selectively emphasize risk cues or benign context to produce misleading conclusions without fabricating evidence.
Each case includes arguments and rebuttals from both sides, supporting evaluation under both one-sided persuasion and competing interpretations.
Text quotations and frame references are checked against the supplied sources to focus evaluation on misleading interpretations of source-grounded evidence.

\section{Problem Definition}
\label{sec:problem}
Let $\mathcal{D}=\{(x_i,y_i,c_i)\}_{i=1}^{N}$ denote MechScam. Each video is represented as $x=(\mathcal{F},\mathcal{S})$, where $\mathcal{F}$ contains identified video frames and $\mathcal{S}$ contains addressable text sources, including the title, description, and transcript segments. The binary label is $y\in\{\textsc{Scam},\textsc{Non-scam}\}$, and the primary category is $c\in\mathcal{C}_{S}\cup\mathcal{C}_{N}$, where $\mathcal{C}_{S}$ and $\mathcal{C}_{N}$ contain six scam mechanisms and six non-scam categories, respectively. The labels are hierarchically consistent: $c\in\mathcal{C}_{S}$ if and only if $y=\textsc{Scam}$.

We consider scam detection, which predicts $y$, and fine-grained classification, which predicts $c$. To evaluate robustness to misleading arguments, we additionally provide a judge with an argument transcript $T$ alongside the original video evidence and ask it to predict the same labels. The transcript contains interpretations that are evaluated against the original video evidence.

Let $J_y(x)$ and $J_y(x,T)$ denote the judge's binary predictions without and with the transcript, respectively. An argument misleads the judge when an initially correct prediction becomes incorrect after the transcript is introduced. We quantify this effect using the misled rate:
\begin{equation}
\mathrm{MR}(J)=
\frac{
\sum_i
\mathbf{1}[J_y(x_i)=y_i]\,
\mathbf{1}[J_y(x_i,T_i)\neq y_i]
}{
\sum_i \mathbf{1}[J_y(x_i)=y_i]
},
\end{equation}
where the sums range over evaluated videos and $T_i$ contains an argument supporting the label opposite to $y_i$. A lower misled rate indicates greater resistance to misleading arguments.

\begin{figure*}[t]
\centering
\includegraphics[width=\textwidth]{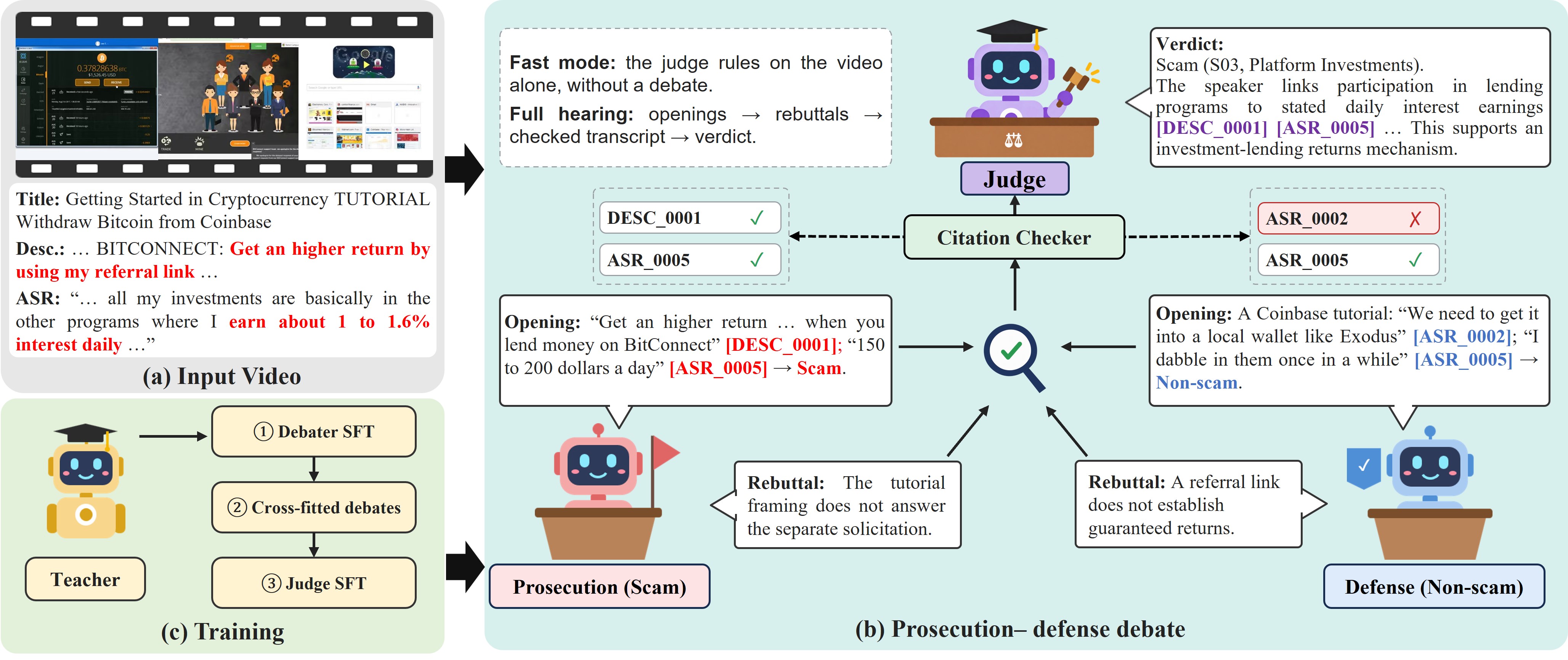}
\caption{Overview of ScamCourt: (a) multimodal video inputs; (b) Prosecution--defense debate with citation verification and judge prediction; and (c) debater training and cross-fitted judge training.}

\label{fig:overview}
\end{figure*}

\section{Methodology}
\label{sec:method}
ScamCourt combines opposing advocacy with source-checked judgment, as illustrated in Figure~\ref{fig:overview}. The prosecution and defense develop competing interpretations of the same video evidence, and a judge evaluates their arguments after citation verification. The framework separates two requirements: citations must refer to the supplied sources, and conclusions must be justified by those sources. We train the debater to sustain both positions and the judge to resolve their disagreement using cross-fitted debates.

\subsection{Verifiable-Citation Debate}
The prosecution argues for Scam and a corresponding scam category, and the defense argues for Non-scam and a corresponding benign category. Both receive the same frames and text sources. Each side first presents an opening and then rebuts the opposing opening, producing a transcript $T=(o_P,o_D,b_P,b_D)$, where $o$ and $b$ denote openings and rebuttals and $P$ and $D$ denote prosecution and defense.

Before judgment, a checker applies $v(e,x)$ to each citation $e$. A text citation is verified when its quotation is a contiguous excerpt of the identified source; a frame citation is verified when its identifier refers to a supplied frame. All other citations are marked unverified. The resulting transcript $\tilde{T}$ retains these verification marks. Verification establishes source provenance. The judge treats the transcript as advocacy and assesses interpretations and visual claims against the original video evidence, returning a binary label, primary category, content role, and supporting explanation.

\subsection{Debater Training}
\label{sec:debater-training}
We fine-tune the backbone (Qwen3-VL-8B-Instruct by default) with a shared, role-conditioned LoRA adapter~\cite{hu2022lora}. Supervision comprises openings and rebuttals for both the reference and opposing positions from the argument set in Section~\ref{sec:adv-set}. For each training video, a zero-shot Qwen3-VL-8B selector scores the candidate debates by the probability assigned to the incorrect label. We retain a top-scoring candidate and its corresponding reference-side statements; selection rules and training counts appear in Appendix~\ref{app:adv-generation}.

The prompt specifies the assigned side and debate stage without revealing whether that side agrees with the reference label. Because the prosecution consistently advocates Scam and the defense Non-scam, the same adapter learns to argue either position without being told which is correct. 

\subsection{Judge Training on Cross-Fitted Debates}
A judge trained only on teacher arguments may encounter a different distribution of arguments when deployed with the learned debater. Generating judge-training debates with a debater fitted to those same videos can further introduce in-sample effects. We address this mismatch through two-fold cross-fitting within the training set. We partition $\mathcal{D}_{tr}$ into $\mathcal{D}_A$ and $\mathcal{D}_B$, stratified by primary category and source collection. A debater trained on $\mathcal{D}_A$ generates debates for $\mathcal{D}_B$, and vice versa, so that each training video receives a debate generated by the debater fitted to the complementary fold. The debater used at test time is subsequently trained on the full training set.

Judge supervision combines teacher-written debates, cross-fitted debates, and transcript-free examples for a stratified 25\% subset of training videos. The first two provide competing interpretations of the evidence, while the third supports judgment without debate. Targets contain the reference label, category, content role, and evidence-grounded rationale. We optimize the answer-token loss, with neither test videos nor their arguments included in training.

\subsection{Inference}

At inference time, the learned debater generates both openings and rebuttals, and the judge evaluates the source-checked transcript. For binary prediction, we normalize the probabilities of the tokens distinguishing the two labels:
\begin{equation}
s_J(x,\tilde{T})=
\frac{p_J(t_{\textsc{Scam}}\mid x,\tilde{T})}
{p_J(t_{\textsc{Scam}}\mid x,\tilde{T})+
 p_J(t_{\textsc{Non-scam}}\mid x,\tilde{T})},
\end{equation}
where $t_{\ell}$ is the first distinguishing token of label $\ell$ at the label prediction position. The resulting score determines the binary decision, while the remaining output fields are decoded greedily. The same judge can operate without a transcript, omitting debater generation for lower-cost assessment.

\section{Experiments}
\label{sec:exp}


We organize the evaluation around three research questions:
\begin{itemize}
  \item \textbf{RQ1} (\S\ref{sec:rq1}): How challenging are scam detection and fine-grained attribution, particularly on benign look-alikes?

  \item \textbf{RQ2} (\S\ref{sec:args}): Does prompted debate or one-sided advocacy improve detection and fine-grained category attribution?

  \item \textbf{RQ3} (\S\ref{sec:rq3}): How challenging are misleading arguments grounded in genuine evidence for baseline scam detectors?
\end{itemize}

\begin{table*}[t]
\caption{Main results (\%). FPR hard is measured on the hard-negative subset. Misled rate is the share of initially correct predictions flipped by a misleading argument under the Single and Full settings. Bold: best in each column.}
\label{tab:main}
\centering
\setlength{\tabcolsep}{4pt}
\begin{tabularx}{\textwidth}{@{}>{\raggedright\arraybackslash}p{0.20\textwidth}*{9}{>{\centering\arraybackslash}X}@{}}
\toprule
 & \multicolumn{7}{c}{Detection} & \multicolumn{2}{c}{Misled rate} \\
\cmidrule(lr){2-8}\cmidrule(l){9-10}
Model & Acc & \mbox{Macro-F1} & \mbox{Scam P} & \mbox{Scam R} & FPR & \mbox{FPR hard} & \mbox{12-way F1} & Single & Full \\
\midrule
\tablegroup{10}{Closed-source MLLMs (zero-shot)}
GPT-6.1-sol & 88.0 & 86.9 & 77.6 & 89.2 & 12.6 & 20.6 & 72.2 & 4.7 & 4.0 \\
Gemini-3.1-Pro & 88.6 & 87.3 & 81.0 & 85.3 & 9.8 & 16.0 & 70.7 & \textbf{0.7} & 0.9 \\
Claude-Opus-5.5 & 89.3 & 87.9 & 83.0 & 84.9 & 8.6 & 14.0 & 67.1 & 0.8 & 1.6 \\
\midrule
\tablegroup{10}{Open-source MLLMs (zero-shot)}
Qwen3-VL-235B & 88.1 & 86.6 & 83.0 & 80.7 & 8.1 & 13.3 & 49.7 & 8.4 & 1.6 \\
GLM-5.3-Flash & 89.1 & 87.6 & 84.4 & 82.1 & 7.4 & 12.0 & 70.2 & 1.2 & 1.0 \\
DeepSeek-V4.1-Flash & 86.7 & 86.2 & 79.5 & 84.4 & 10.7 & 17.3 & 70.8 & 3.0 & 2.7 \\
\midrule
\tablegroup{10}{ScamCourt}
Qwen3-VL (8B) & 82.3 & 81.4 & 67.1 & 90.8 & 21.8 & 34.8 & 27.7 & 8.5 & 4.1 \\
\quad w/ self-debate & 80.9 & 80.1 & 64.8 & \textbf{92.0} & 24.5 & 39.6 & 39.3 & -- & -- \\
\quad w/ SFT & 92.7 & 91.7 & 90.3 & 87.4 & 4.6 & 7.6 & 75.8 & 12.9 & 3.3 \\
\quad w/ ScamCourt & 93.4 & 92.3 & \textbf{93.9} & 85.3 & \textbf{2.7} & \textbf{4.4} & 77.1 & 2.2 & \textbf{0.2} \\
GLM-4.1V (9B) & 83.8 & 81.6 & 76.2 & 74.1 & 11.4 & 18.4 & 40.7 & 57.6 & 10.5 \\
\quad w/ self-debate & 79.2 & 77.6 & 65.3 & 78.9 & 20.6 & 30.0 & 37.5 & -- & -- \\
\quad w/ SFT & 92.9 & 91.8 & 92.1 & 85.8 & 3.6 & 5.9 & 76.2 & 12.2 & 2.3 \\
\quad w/ ScamCourt & \textbf{94.4} & \textbf{93.6} & 92.9 & 89.9 & 3.4 & 5.2 & \textbf{79.1} & 4.4 & \textbf{0.2} \\
\bottomrule
\end{tabularx}
\end{table*}

\subsection{Experimental Settings}
\paragraph{Data and metrics.}
\label{sec:metrics}
We use a frozen split of 5,290 training and 1,324 test videos, constructed with seed 42 using category--source strata and retained after label review. The test set contains 436 scam and 888 non-scam videos; 543 non-scam videos contain candidate risk elements and form the hard-negative subset. We report binary accuracy, macro-F1, scam precision and recall, 12-way macro-F1, and False-Positive Rates (FPR) over all non-scam videos and hard negatives. Robustness to misleading arguments uses the misled rate in Section~\ref{sec:problem}, conditioned on each judge's initially correct predictions.

\paragraph{Models and inputs.}
\label{sec:impl}
We evaluate GPT-6.1-sol, Gemini-3.1-Pro, Claude-Opus-5.5, Qwen3-VL-235B-A22B-Instruct, GLM-5.3-Flash, and DeepSeek-V4.1-Flash without task-specific fine-tuning. ScamCourt is instantiated on two backbones, Qwen3-VL-8B-Instruct~\cite{qwen3vl} and GLM-4.1V-9B-Thinking~\cite{glm41v}. For each backbone, we also report zero-shot judgment, prompted self-debate, and a supervised classifier (SFT) trained on the same videos without debates. All models receive twelve frames with a per-frame pixel budget of $384^2$, the title, uploader-provided description, and ASR. Predictions contain the label, category, content role, explanation, and citations. Fine-tuned variants share LoRA and optimization settings; the debater trains for one epoch and the judge for two. Further details appear in Appendix~\ref{app:implementation}.

\paragraph{Misleading-argument evaluation.}
\label{sec:adv-eval}
The misleading argument test set contains three arguments per test video, totaling 3,972 instances. Alongside the video evidence, judges receive either the misleading opening alone (\emph{Single}) or both openings and rebuttals in prosecution--defense order (\emph{Full}). A fixed zero-shot Qwen3-VL-8B selector chooses the argument with the highest wrong-label probability per video; all judges receive the same selected misleading arguments. Sample-specific inference inputs comprise only the video evidence and supplied arguments, and training uses arguments drawn exclusively from training videos.

\subsection{Main Results (RQ1)}
\label{sec:rq1}
We examine scam detection and fine-grained category attribution using the results in Table~\ref{tab:main} and per-category scores in Table~\ref{tab:perclass}, focusing on benign look-alikes and category boundaries.

\begin{table*}[t]
\centering
\caption{Per-category and binary F1 (\%) on the test set. S01--N06 report 12-way class-specific F1; Scam and Non-scam report binary class-specific F1. Bold: best value in each column.}
\label{tab:perclass}
\setlength{\tabcolsep}{0.5pt}
\begin{tabularx}{\textwidth}{@{}>{\raggedright\arraybackslash}p{0.18\textwidth}*{14}{>{\centering\arraybackslash}X}@{}}
\toprule
 & \multicolumn{12}{c}{Fine-grained F1} & \multicolumn{2}{c}{Binary F1} \\
\cmidrule(lr){2-13}\cmidrule(l){14-15}
Model & S01 & S02 & S03 & S04 & S05 & S06 & N01 & N02 & N03 & N04 & N05 & N06 & \makebox[\linewidth][l]{Scam} & \makebox[\linewidth][r]{Non-scam} \\
\midrule
\tablegroup{15}{Closed-source MLLMs (zero-shot)}
GPT-6.1-sol & 42.2 & 78.5 & 67.8 & 84.2 & 42.3 & 82.9 & \textbf{92.2} & 82.6 & 76.5 & 66.3 & 69.2 & 82.2 & 83.0 & 90.7 \\
Gemini-3.1-Pro & 39.7 & 78.3 & 56.9 & 77.4 & 45.0 & 78.3 & 88.7 & 88.8 & 77.6 & 63.2 & 69.9 & 84.8 & 83.1 & 91.4 \\
Claude-Opus-5.5 & 43.9 & 73.8 & 57.4 & 68.8 & 35.8 & 76.7 & 88.9 & 84.5 & 73.0 & 59.3 & 65.5 & 77.5 & 83.9 & 92.0 \\
\midrule
\tablegroup{15}{Open-source MLLMs (zero-shot)}
Qwen3-VL-235B & 27.6 & 53.1 & 31.8 & 40.7 & 33.7 & 70.2 & 60.6 & 58.6 & 61.8 & 57.7 & 51.0 & 49.4 & 81.9 & 91.3 \\
GLM-5.3-Flash & 42.9 & 77.6 & 57.1 & 80.9 & 41.3 & 80.2 & 91.2 & 87.2 & 74.5 & 65.3 & 65.6 & 78.3 & 83.3 & 91.9 \\
\mbox{DeepSeek-V4.1-Flash} & 45.0 & 81.3 & 64.2 & 78.3 & 38.8 & 81.7 & 90.7 & 83.7 & 75.7 & 68.0 & 64.5 & 77.5 & 81.9 & 90.6 \\
\midrule
\tablegroup{15}{ScamCourt}
Qwen3-VL (8B) & 22.1 & 35.0 & 20.8 & 19.2 & 14.6 & 50.8 & 41.3 & 35.5 & 52.8 & 22.0 & 16.9 & 1.6 & 77.2 & 85.6 \\
\quad w/ self-debate & 15.5 & 74.8 & 30.6 & 20.4 & 3.1 & 57.1 & 57.9 & 60.1 & 62.6 & 17.1 & 24.4 & 48.0 & 76.0 & 84.1 \\
\quad w/ SFT & 51.4 & 81.8 & 66.7 & 82.6 & 47.7 & 80.8 & 83.9 & 92.1 & 82.1 & 72.9 & 80.3 & 87.1 & 88.8 & 94.6 \\
\quad w/ ScamCourt & 56.8 & 82.7 & \textbf{70.9} & \textbf{85.2} & 49.6 & \textbf{83.3} & 86.2 & 91.0 & 81.6 & 75.4 & 76.8 & 85.7 & 89.4 & 95.2 \\
GLM-4.1V (9B) & 17.9 & 32.7 & 23.0 & 0.0 & 24.5 & 64.2 & 42.4 & 75.2 & 47.3 & 45.8 & 51.6 & 64.3 & 75.1 & 88.0 \\
\quad w/ self-debate & 21.2 & 55.0 & 29.1 & 16.7 & 32.1 & 45.9 & 25.7 & 66.7 & 43.0 & 30.1 & 31.8 & 53.3 & 71.4 & 83.7 \\
\quad w/ SFT & 51.4 & \textbf{86.5} & 65.6 & 82.2 & 43.8 & 83.1 & 86.2 & 92.7 & 82.8 & 73.9 & 80.6 & 85.6 & 88.8 & 94.8 \\
\quad w/ ScamCourt & \textbf{62.5} & 85.5 & 69.6 & 83.2 & \textbf{50.0} & 82.4 & 86.7 & \textbf{93.0} & \textbf{83.8} & \textbf{76.5} & \textbf{87.3} & \textbf{88.4} & \textbf{91.4} & \textbf{95.9} \\
\bottomrule
\end{tabularx}
\end{table*}

\begin{table}[t]
\caption{Self-debate and one-sided advocacy with MLLMs (\%).}
\label{tab:selfdebate}
\setlength{\tabcolsep}{2pt}
\resizebox{\columnwidth}{!}{%
\begin{tabular}{@{}llccccc@{}}
\toprule
Model & Protocol & Acc & Macro-F1 & FPR & FPR hard & 12-way F1 \\
\midrule
GPT-6.1 & Self-debate & 87.0 & 85.6 & 11.6 & 19.0 & 71.6 \\
 & Prosecution only & 86.0 & 84.8 & 14.5 & 23.8 & 70.4 \\
 & Defense only & 86.5 & 85.3 & 14.2 & 23.2 & 72.0 \\
\midrule
Gemini & Self-debate & 88.8 & 87.5 & 9.9 & 16.2 & \textbf{74.2} \\
 & Prosecution only & 88.9 & 87.6 & 9.3 & 15.3 & 72.8 \\
 & Defense only & 89.0 & \textbf{87.9} & 11.0 & 18.0 & 72.5 \\
\midrule
Claude & Self-debate & 88.8 & 87.5 & 9.2 & 14.7 & 72.7 \\
 & Prosecution only & \textbf{89.1} & 87.8 & 8.8 & 14.2 & 71.5 \\
 & Defense only & 88.4 & 87.0 & 9.6 & 15.3 & 71.3 \\
\midrule
Qwen3-VL-235B & Self-debate & 85.9 & 84.0 & 9.7 & 15.8 & 59.4 \\
 & Prosecution only & 87.4 & 86.0 & 10.1 & 16.4 & 50.3 \\
 & Defense only & 86.9 & 84.5 & \textbf{4.6} & \textbf{7.6} & 56.4 \\
\midrule
GLM-5.3 & Self-debate & 88.7 & 87.2 & 8.8 & 14.4 & 72.0 \\
 & Prosecution only & 88.1 & 86.6 & 9.1 & 14.7 & 66.8 \\
 & Defense only & \textbf{89.1} & 87.6 & 7.5 & 12.3 & 70.6 \\
\midrule
DeepSeek-V4.1 & Self-debate & 86.8 & 86.0 & 11.8 & 19.3 & 72.0 \\
 & Prosecution only & 87.0 & 86.0 & 12.3 & 20.1 & 69.9 \\
 & Defense only & 88.1 & 87.3 & 10.6 & 17.3 & 72.5 \\
\bottomrule
\end{tabular}}
\end{table}

\paragraph{Current MLLMs.}
As shown in Table~\ref{tab:main}, binary detection and mechanism attribution exhibit different performance patterns. Qwen3-VL-235B combines competitive binary accuracy with the lowest 12-way macro-F1 among the larger MLLMs, highlighting the importance of evaluating both tasks.
Across the larger MLLMs, false-positive rates are higher on hard negatives than on the full benign set, identifying benign videos with risk cues as a central challenge for contextual discrimination.

\paragraph{Category-level challenges.}
As shown in Table~\ref{tab:perclass}, Reward Claims (S01) and Token Investments (S05) are particularly challenging for the larger MLLMs, with F1 ranging from 27.6\% to 45.0\% and from 33.7\% to 45.0\%, respectively.
For example, Gemini and Claude classify an iTunes credit-redemption tutorial (N05) as S01, illustrating how shared reward--action cues can obscure an ordinary promotional mechanism.
For S05, a Luna video combines price commentary with a purchase invitation and a purported zero-risk offer, yet GPT, Gemini, and Claude assign it N02 (Financial News and Analysis).
ScamCourt identifies S05 by linking token purchases to potential appreciation and airdrop eligibility, highlighting the need to connect financial claims with the participation they solicit.
Across its two backbones, ScamCourt achieves the highest F1 in ten of the twelve categories, including five scam mechanisms, with its best S01 and S05 scores reaching 62.5\% and 50.0\%.
S01 and S05 remain challenging after debate training.
On benign content, gains in N04 and N05 complement the hard-negative results. Several larger models perform particularly well on N01.

\paragraph{ScamCourt versus the same-backbone classifier.}
On both backbones, ScamCourt rejects more benign look-alikes and improves 12-way macro-F1 over the same-backbone classifier, showing consistent benefits across model families.
The accompanying category gains support improved discrimination among scam mechanisms and benign contexts, consistent with ScamCourt's emphasis on interpreting promise--action relationships.

\begin{table}[t]
\caption{Ablation studies on the Qwen3-VL (8B) backbone (\%).}
\label{tab:ablation}
\setlength{\tabcolsep}{4pt}
\begin{tabular}{@{}lccc@{}}
\toprule
Variant & Acc & Macro-F1 & FPR hard \\
\midrule
ScamCourt & \textbf{93.4} & \textbf{92.3} & \textbf{4.4} \\
\quad w/o debate at inference & 93.3 & \textbf{92.3} & 6.1 \\
\quad w/o debate training & 92.7 & 91.7 & 7.6 \\
\quad w/o any training & 82.3 & 81.4 & 34.8 \\
\bottomrule
\end{tabular}
\end{table}

\begin{figure*}[t]
\centering
\includegraphics[width=\textwidth]{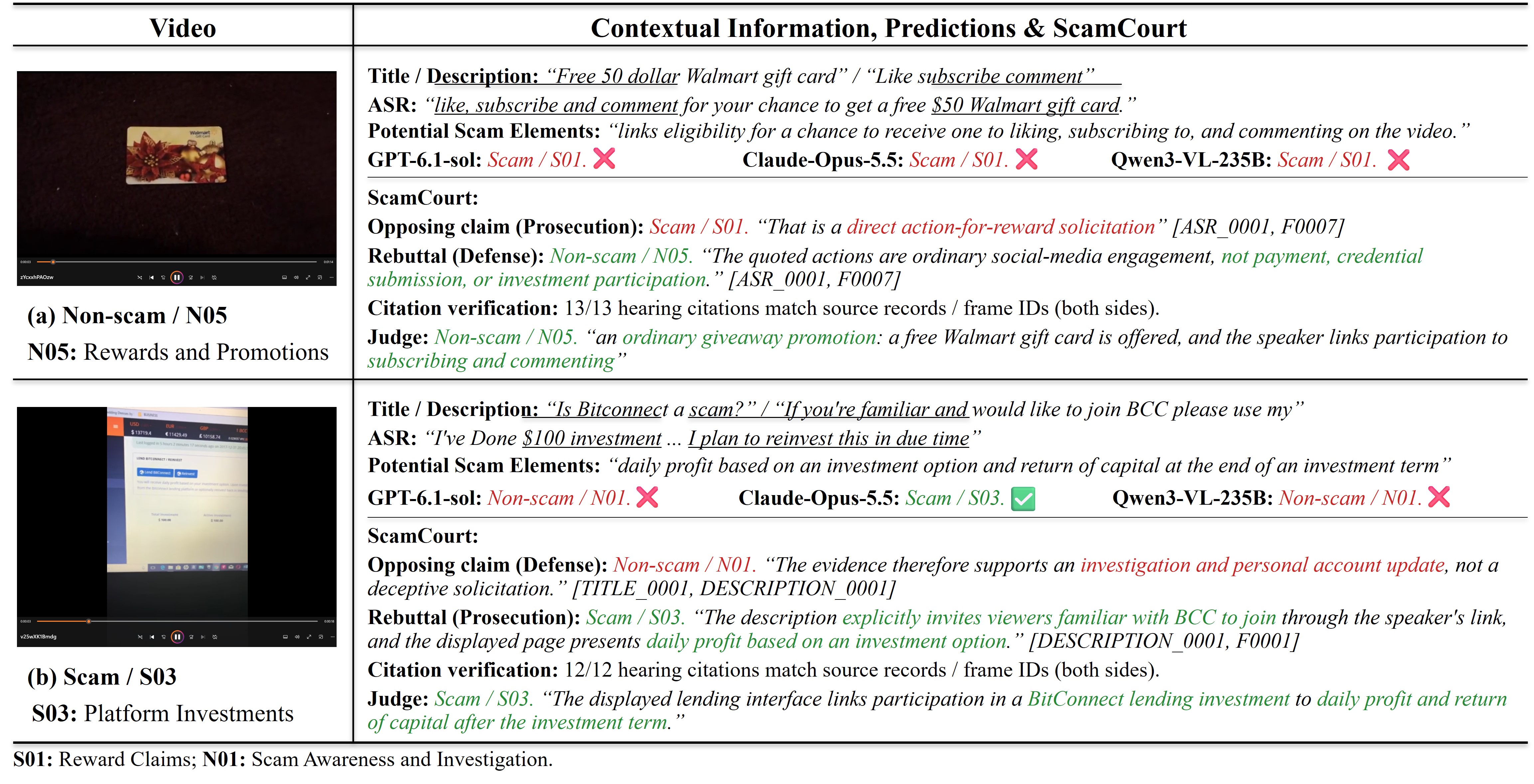}
\caption{The test cases comparing MLLM predictions with ScamCourt. Each case illustrates ScamCourt's opposing claims, rebuttals, citation verification, and final verdict.}
\label{fig:case-study}
\end{figure*}

\subsection{Effects of Debate and One-Sided Advocacy (RQ2)}
\label{sec:args}
We assess whether prompted self-debate and one-sided advocacy improve scam detection and fine-grained category attribution over the zero-shot predictions in Table~\ref{tab:main}. Self-debate includes both openings and rebuttals. One-sided advocacy follows consultancy~\cite{khan2024debating}, providing only a prosecution or defense opening. All arguments undergo citation checks.

As shown in Table~\ref{tab:selfdebate}, prompted debate affects binary judgment and category attribution differently. Self-debate improves 12-way macro-F1 for most larger MLLMs, with model-dependent changes in binary accuracy and hard-negative false positives. These patterns suggest that articulating competing interpretations aids category attribution; reliable binary decisions require effective evaluation of those interpretations. One-sided advocacy also produces stance-dependent shifts. For Qwen3-VL-235B, defense-only input reduces both false positives and overall accuracy; prosecution-only input increases false positives. These findings motivate explicit supervision of the judge's argument assessment.

\subsection{Challenges Posed by Misleading Arguments (RQ3)}
\label{sec:rq3}
We assess how misleading arguments grounded in genuine video evidence challenge scam detection through the misled-rate columns in Table~\ref{tab:main}.

\paragraph{Challenges for baseline detectors.}
Baseline detectors differ substantially in their susceptibility. Under \emph{Single}, zero-shot GLM-4.1V is readily swayed, while the Qwen3-VL SFT classifier is more often misled than its zero-shot counterpart despite stronger detection accuracy. These failures concern interpretations of genuine evidence and support evaluating resistance to misleading arguments as a separate capability.

\paragraph{Effects of competing arguments.}
The effects of providing both sides vary across models, with a substantial benefit for Qwen3-VL-235B. Under the same misleading arguments, ScamCourt outperforms the zero-shot and SFT variants on both backbones in both settings. Several larger models remain more robust under \emph{Single}. ScamCourt achieves the lowest misled rate under \emph{Full}. These results highlight the challenge of resolving competing interpretations and support evaluating resistance to misleading arguments separately from detection accuracy.

\subsection{Ablation Study}
\label{sec:ablation}

Table~\ref{tab:ablation} evaluates debate training and debate at inference through progressively simplified variants on the Qwen3-VL (8B) backbone. In transcript-free inference, the debate-trained judge outperforms the classifier in accuracy and hard-negative rejection, supporting the transfer of debate supervision to the judge's own predictions. Providing a generated debate further reduces hard-negative false positives while maintaining similar overall accuracy. These comparisons show that debate supervision benefits transcript-free detection and that debate at inference further improves contextual discrimination.

\subsection{Case Study}
Figure~\ref{fig:case-study} presents two examples of ScamCourt's judgments.

In case (a), all three baselines classify a \$50 gift-card giveaway as Scam/S01.
Both S01 and N05 can link rewards to participation, making the requested action and its context crucial to distinguishing them.
The prosecution treats the like--subscribe--comment requirement as a deceptive reward mechanism, while the defense identifies it as ordinary audience engagement without financial or sensitive-data solicitation.
The judge returns Non-scam/N05 (Rewards and Promotions), preserving the distinction between promotional participation and deceptive reward acquisition.

In case (b), two baselines assign Non-scam/N01 to \emph{Is Bitconnect a scam?}. Claude-Opus-5.5 correctly predicts Scam/S03.
The prosecution counters the investigative framing by connecting the description's invitation to join with the lending interface's daily-profit and capital-return claims.
The judge returns Scam/S03 (Platform Investments), identifying a deposit-for-return solicitation beneath the questioning title.
The distinction from N01 lies in the creator's role in directing viewers toward the investment, which requires interpreting the invitation alongside the displayed return mechanism.
Both sides' citations pass source checks in both cases. Correct attribution depends on the relationship between the cited evidence, communicative context, and proposed category.

\section{Conclusion}
In this study, we highlight the importance of mechanism-level reasoning for distinguishing scam videos from non-scam videos with similar risk cues. To this end, we introduce MechScam, a human-verified benchmark spanning six scam mechanisms and six benign categories with evidence-grounded rationales and misleading arguments, and ScamCourt, a verifiable-citation debate framework in which trained debaters argue over shared evidence and a trained judge rules on the source-checked transcript. Extensive experiments show that ScamCourt outperforms larger MLLMs and same-backbone baselines across two model families, rejecting more benign look-alikes and resisting evidence-grounded misleading arguments. Together, these contributions provide a foundation for mechanism-aware, interpretable, and robust scam video detection on the Web.

\bibliographystyle{ACM-Reference-Format}
\bibliography{references}

\appendix

\section{Illustrative Category Examples}
\label{app:category-examples}
The following training samples illustrate all twelve categories. Original frames provide visual context, and descriptions summarize the video's multimodal evidence.

\newcommand{\categoryexample}[6]{%
  \par\noindent
  \begingroup
  \small\setlength{\tabcolsep}{3pt}%
  \begin{tabular}{@{}>{\raggedright\arraybackslash}p{0.29\columnwidth}>{\raggedright\arraybackslash}p{\dimexpr0.71\columnwidth-6pt\relax}@{}}
  \toprule
  \begin{minipage}[t]{\linewidth}
    \vspace{0pt}\centering
    #6\par
    {\footnotesize\texttt{\detokenize{#2}}\par #3}
  \end{minipage}
  & \textbf{#1: #4}\par #5 \\
  \end{tabular}
  \endgroup\par}

\subsection{Scam Mechanisms}
\categoryexample{S01}{3BgZ3x-4AL0}{F0012}{Reward Claims}{An external Gardenscapes generator promises free coins and requires installing two apps, linking the reward to a claim workflow.}{\includegraphics[width=\linewidth,height=32pt,keepaspectratio]{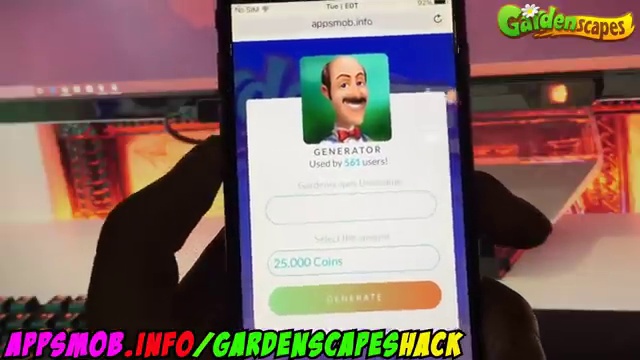}}

\categoryexample{S02}{30TD7exZIOE}{F0041}{Trading Bots}{A BNB bot tutorial asks viewers to fund a deployed contract and activate it for purported trading profits.}{\includegraphics[width=\linewidth,height=32pt,keepaspectratio]{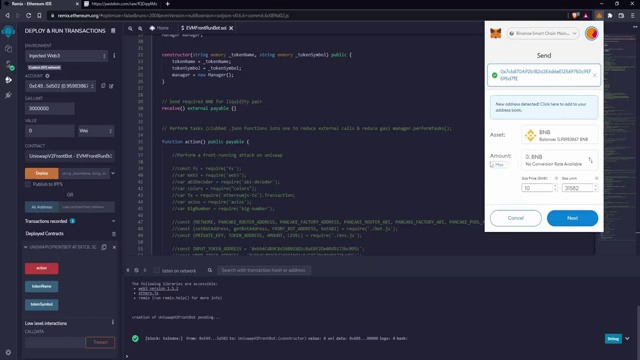}}

\categoryexample{S03}{-0Gsg1NySzU}{F0001}{Platform Investments}{BitConnect lending and reinvestment are promoted as a route to daily income, with returns attributed to deposited funds.}{\includegraphics[width=\linewidth,height=32pt,keepaspectratio]{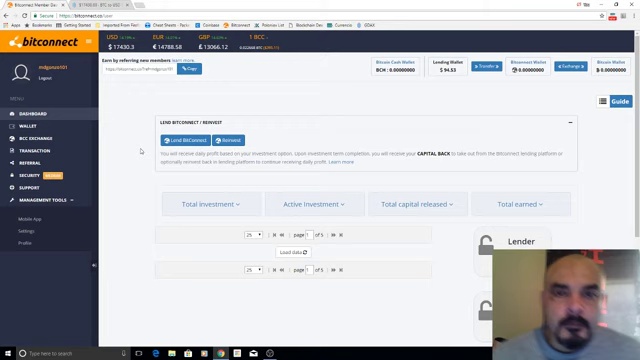}}

\categoryexample{S04}{-IBWtxrbDaA}{F0041}{Mining and Nodes}{A \$500 BitClub mining purchase is linked to daily Bitcoin earnings, making paid mining resources the central return mechanism.}{\includegraphics[width=\linewidth,height=32pt,keepaspectratio]{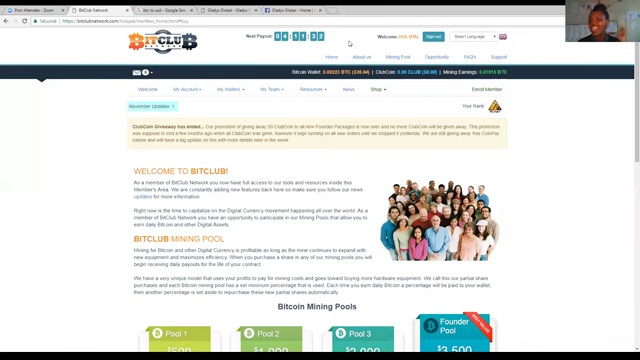}}

\categoryexample{S05}{-VHf0hhzxaE}{F0012}{Token Investments}{Polybius ICO tokens are promoted through potential resale gains and dividends, placing token ownership at the center of the offer.}{\includegraphics[width=\linewidth,height=32pt,keepaspectratio]{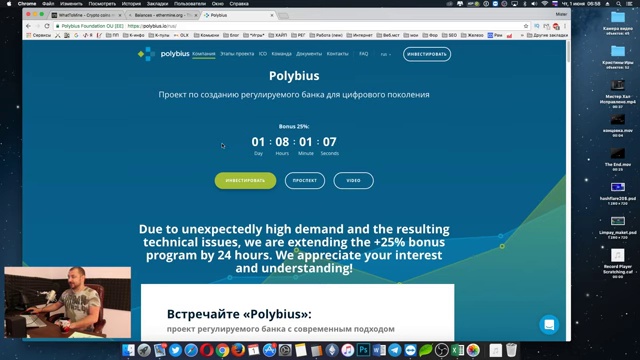}}

\categoryexample{S06}{--WbWqCAApA}{F0030}{Recruitment Commissions}{Forsage rewards depend on invited partners filling matrix positions, with multiple paid levels and transfers to an upline.}{\includegraphics[width=\linewidth,height=32pt,keepaspectratio]{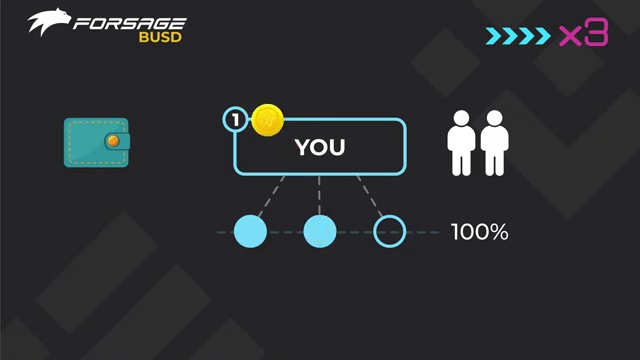}}

\subsection{Non-scam Contexts}
\categoryexample{N01}{2pYQdcB5ZLU}{F0007}{Scam Awareness and Investigation}{An anti-scam caller tests purported Microsoft support; the tested party requests payment during the scambaiting interaction.}{\includegraphics[width=\linewidth,height=32pt,keepaspectratio]{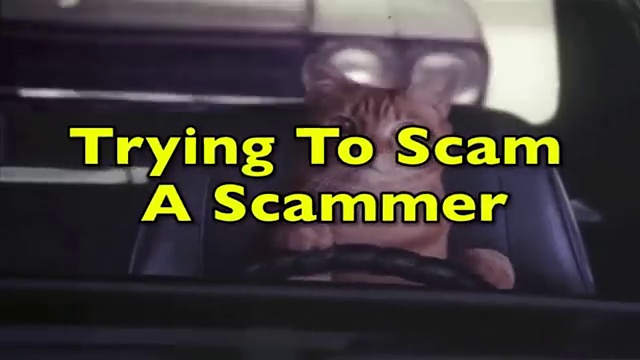}}

\categoryexample{N02}{7IkcNOTE20s}{F0012}{Financial News and Analysis}{The presenter evaluates a Goldman Sachs stock portfolio and its holdings, focusing on market commentary.}{\includegraphics[width=\linewidth,height=32pt,keepaspectratio]{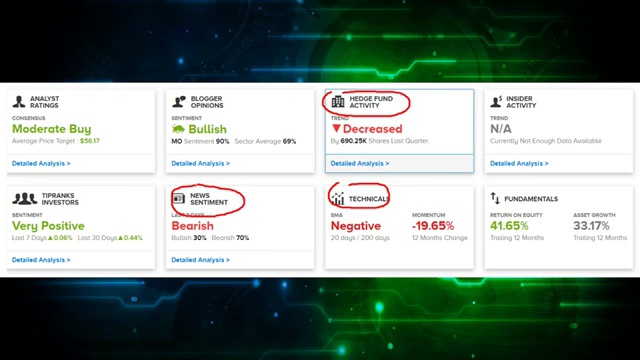}}

\categoryexample{N03}{dSepj_pU4Rk}{F0018}{Education and Tutorials}{An open-source Uniswap Python development session shows code and tests, using cryptocurrency infrastructure as a programming subject.}{\includegraphics[width=\linewidth,height=32pt,keepaspectratio]{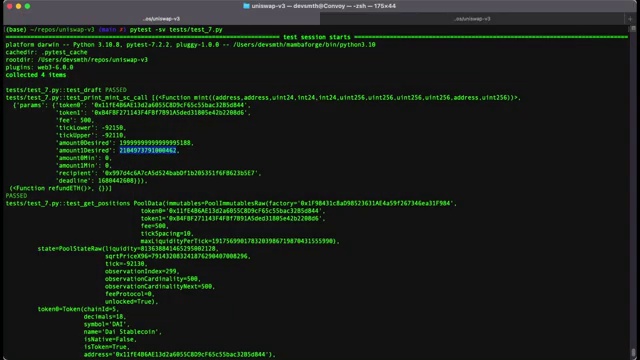}}

\categoryexample{N04}{0EJX5CH2sGs}{F0001}{Products and Services}{A QVC presentation sells an HP laptop with a displayed price and included support, illustrating ordinary retail commerce.}{\includegraphics[width=\linewidth,height=32pt,keepaspectratio]{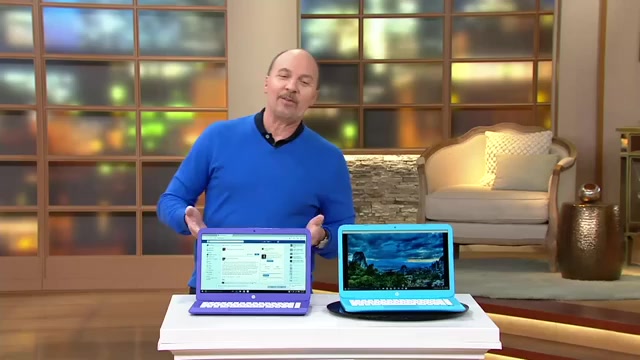}}

\categoryexample{N05}{1oufYWwgDcs}{F0047}{Rewards and Promotions}{Microsoft Rewards points earned through searches and quizzes are redeemed for Xbox membership, illustrating an ordinary loyalty scheme.}{\includegraphics[width=\linewidth,height=32pt,keepaspectratio]{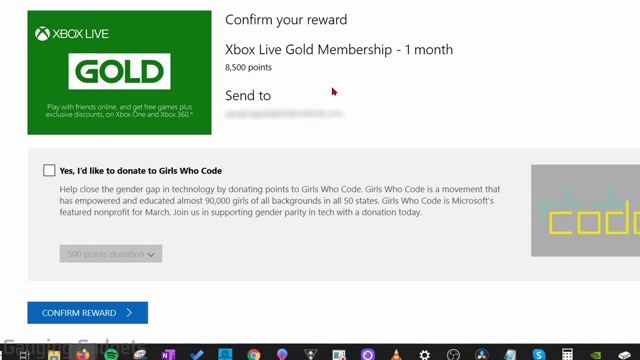}}

\categoryexample{N06}{0AZ26cIz4MQ}{F0047}{General Content}{A masala pav recipe demonstrates preparation and presents the finished dish, illustrating general food and lifestyle content.}{\includegraphics[width=\linewidth,height=32pt,keepaspectratio]{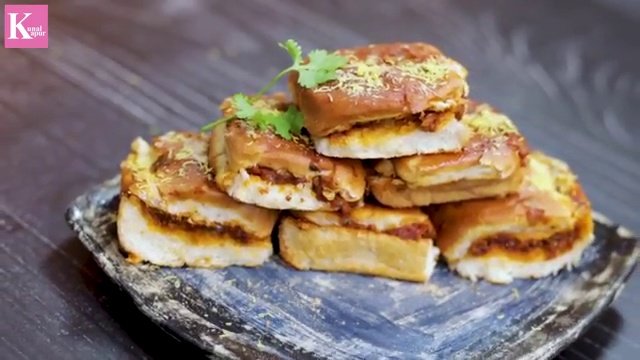}}

\section{Annotation Verification and Label Corrections}
\label{app:quality-control}

\paragraph{Review procedure.}
Automatic checks verify output structure, label--category compatibility, verbatim text quotations, and frame identifiers, establishing citation provenance.
Two computer science master's students working on anti-fraud research independently review every video's label, category, and rationale against the original material.
Rejected or disputed cases are adjudicated by a four-member panel comprising these reviewers and two doctoral students in the same field.
During re-annotation, prior labels and rationales are withheld to support independent reassessment of the source material.

\subsection{Label Corrections}
\label{app:relabel}

We correct 124 source labels (1.9\% of the corpus), mostly to distinguish anti-scam warnings and investigations from the schemes they describe.
Table~\ref{tab:relabel} reports the corrections by collection, counting a correction whenever the final binary label differs from the source label.

\begin{table}[H]
\caption{Source-label statistics and retained binary-label corrections in MechScam. Corrected counts final labels that differ from source labels.}
\label{tab:relabel}
\small
\setlength{\tabcolsep}{4pt}
\begin{tabular}{@{}llrrr@{}}
\toprule
Collection & Subset & Videos & Source scam & Corrected \\
\midrule
VidScam & Gift card & 1,878 & 86 & 10 \\
 & Monetary & 789 & 76 & 3 \\
 & Cryptocurrency & 412 & 378 & 4 \\
YouTube\_Scam & Scam cases & 1,722 & 1,722 & 90 \\
 & Legitimate crypto & 1,813 & 0 & 17 \\
\midrule
Total & & 6,614 & 2,262 & 124 \\
\bottomrule
\end{tabular}
\end{table}

\section{Implementation and Prompt Details}
\label{app:implementation}

\subsection{Misleading Argument Construction}
\label{app:adv-generation}

\paragraph{Generation and grounding.}
Openings contain at most 180 words and one to five citations; rebuttals and reference-side replies contain at most 120 words and up to four citations.
Arguments selectively interpret genuine content under constraints against fabricating facts, actions, or outcomes.
Text quotations are checked against named sources, and visual references must identify supplied frames.
Reference-side openings are rewritten in a comparable advocacy style under the same grounding constraints.
When strong misleading arguments cannot be grounded, weaker cases are retained under the same source constraints.

\paragraph{Selection and supervision.}
A fixed zero-shot Qwen3-VL-8B selector ranks candidate transcripts by the probability assigned to the incorrect label.
For training, scores within $10^{-3}$ of the maximum are treated as tied and resolved by a deterministic video-identifier hash; test selection uses the strict maximum and retains the first candidate in an exact tie.
All three candidates remain available.
After filtering, supervision contains 5,269 misleading openings, 5,270 misleading rebuttals, 5,258 reference-side openings, and 5,266 reference-side replies, all from training videos.

\subsection{Model Settings}

\paragraph{Generation and parsing.}
The shared text budget prioritizes titles and descriptions while distributing transcript excerpts across each video.
Larger MLLMs use temperature 0 and at most 1,200 output tokens; local debaters generate at most 512 new tokens, with up to three attempts to obtain a valid statement.
Failed statements are explicitly marked, and outputs without a recoverable label count as errors.
The same parsing rules recover recognized category codes and explicit labels preceding truncation across models.
For local judges, binary probabilities are read at the first differing token of the two label continuations.

\paragraph{Optimization.}
Fine-tuned variants use LoRA with rank 16, $\alpha=32$, and dropout 0.05 on language-model attention and MLP projections.
Training uses a learning rate of $10^{-4}$, 3\% warmup, effective batch size 16, bfloat16, and answer-only loss; sequences exceeding 16,384 tokens are skipped.
Debaters train for one epoch and judges for two; debater training takes approximately 14 hours on one 32\,GB RTX 5090.
For GLM-4.1V-9B-Thinking, LoRA excludes the vision encoder, and zero-shot generation starts after an empty reasoning block.

\subsection{Prompts and Output Formats}
\label{app:prompts}

Tables~\ref{tab:prompts-court}--\ref{tab:prompts-teacher} summarize the debater, judge, and teacher prompts.
All roles treat source content as untrusted data and distinguish depicted schemes from creator promotion.
Text citations contain a source identifier and a contiguous quote of at most 40 words; frame citations contain a supplied frame identifier.

\begin{table}[H]
\caption{Prompt and output format of the ScamCourt debater.}
\label{tab:prompts-court}
\small
\begin{tabular}{@{}>{\raggedright\arraybackslash}p{\columnwidth}@{}}
\toprule
\textbf{Role 1: Debater (prosecution or defense; one shared model)} \\
\textbf{Definition:} Given a video and an assigned side, argue for that side's label and name the category that best fits, citing only evidence that can be checked against the video. \\
\emph{Instruction:} ``You are the PROSECUTION [DEFENSE] in a content-review hearing about ONE video. Argue that the video is Scam [Non-scam] content and name the single scam mechanism [Non-scam context] that best fits. A judge will read both sides and decide. Use only the supplied frames and text sources. Every factual statement about the video must be supported by citations \ldots{} Never invent content. Do not put category codes in your prose.'' The six categories of the assigned side follow. \\
\textbf{Input:} Twelve frames with identifiers; title, description and transcript segments with source identifiers; the stage. At the rebuttal stage, the debater's own opening and the opponent's opening are added. \\
\textbf{Output:} Opening: \texttt{\{"claim\_label":"...",\allowbreak"claimed\_subtype":"<ID>",\allowbreak"argument":"...",\allowbreak"citations":\allowbreak[\{"source\_id":"...",\allowbreak"quote":"..."\}]\}} with at most 180 words and 1--5 citations. Rebuttal: \texttt{\{"rebuttal":"...",\allowbreak"citations":\allowbreak[...]\}} with at most 120 words and 0--4 citations. \\
\bottomrule
\end{tabular}
\end{table}

\begin{table}[H]
\caption{Prompt and output format of the ScamCourt judge.}
\label{tab:prompts-judge}
\small
\begin{tabular}{@{}>{\raggedright\arraybackslash}p{\columnwidth}@{}}
\toprule
\textbf{Role 2: Judge} \\
\textbf{Definition:} Given a video and, optionally, a checked debate transcript, decide the binary label and the category. \\
\emph{Instruction:} ``You are the JUDGE in a content-review hearing about ONE video. Decide whether the video itself is Scam content (promoting or facilitating a suspicious reward, return, recruitment, payment or asset-acquisition solicitation linked to an audience action) or Non-scam content (warning, investigation, news, commentary, education, ordinary commerce, entertainment or other content not promoting a scam mechanism). The frames and text sources are the primary evidence; the debate transcript is advocacy. A [verified] mark only means the quote appears verbatim in the cited source; it does not mean the advocate's interpretation is correct.'' The twelve categories follow. \\
\textbf{Input:} The same frames and text sources, followed by the transcript with a verification mark on every citation, or by the note ``No debate transcript (decide from the video inputs alone).'' \\
\textbf{Output:} \texttt{\{"label":"Scam|Non-scam",\allowbreak"subtype":"<ID>",\allowbreak"content\_role":"...",\allowbreak"verdict":"...",\allowbreak"citations":\allowbreak[...]\}}. The scam probability is read from the first label token. \\
\bottomrule
\end{tabular}
\end{table}

\begin{table}[H]
\caption{Abridged teacher prompts for misleading arguments and reference-side openings.}
\label{tab:prompts-teacher}
\small
\begin{tabular}{@{}>{\raggedright\arraybackslash}p{\columnwidth}@{}}
\toprule
\textbf{Teacher: Misleading arguments} \\
\textbf{Input:} Frames, text sources, reference label and opening, category definitions, and three argument strategies. \\
\emph{Instruction (abridged):} Construct three persuasive cases for the opposite label using distinct strategies and genuine content. Do not invent facts or outcomes, or reveal which side is correct. \\
\textbf{Output:} Three argument records containing a strategy, claimed category, opening, rebuttal, reference-side reply, and source citations. \\
\midrule
\textbf{Teacher: Reference-side openings} \\
\textbf{Input:} Reference rationale and evidence, assigned side, and a misleading opening as a style example. \\
\emph{Instruction (abridged):} Rewrite the reference rationale as an opening in a comparable advocacy style. Preserve its position, claims, and evidence without adding facts. \\
\textbf{Output:} An opening containing an argument and one to five citations to the supplied evidence. \\
\bottomrule
\end{tabular}
\end{table}

\section{Evaluation of Explainability}
\label{app:explainability}

Following the explanation-evaluation framework of \citet{wang2023evaluating}, we assess generated rationales through \emph{Informativeness} (relevant explanatory detail), \emph{Readability} (clarity and fluency), \emph{Soundness} (logical and evidential support), and \emph{Persuasiveness} (convincing justification).
With seed 42, stratified sampling selects 200 test videos, comprising 100 scam and 100 non-scam videos and 16--17 videos per category.
The same videos support paired comparisons of Zero-shot, SFT, and ScamCourt on both backbones.
Under this protocol, GPT-5.5 and two independent human annotators assign scores from 1 to 5 using the same frames, title, description, transcript, and reference annotations.
Only final explanations and their citations are assessed; model identities are hidden, presentation order is randomized, and human scores are averaged.

\begin{table}[H]
\centering
\caption{Rationale quality (1--5; higher is better).}
\label{tab:explainability}
\small
\setlength{\tabcolsep}{0.5pt}
\renewcommand{\arraystretch}{1.08}
\begin{tabularx}{\columnwidth}{@{}>{\raggedright\arraybackslash}p{0.11\columnwidth}*{6}{>{\centering\arraybackslash}X}@{}}
\toprule
 & \multicolumn{3}{c}{GPT-5.5} & \multicolumn{3}{c}{Human} \\
\cmidrule(lr){2-4}\cmidrule(l){5-7}
Metric & \mbox{Zero-shot} & SFT & \mbox{ScamCourt} & \mbox{Zero-shot} & SFT & \mbox{ScamCourt} \\
\midrule
\tablegroup{7}{Qwen3-VL (8B)}
Info. & 3.69 & 3.89 & 3.95 & 3.74 & 3.92 & 4.16 \\
Read. & 4.86 & 4.93 & 4.95 & 4.87 & 4.95 & 4.97 \\
Sound. & 2.90 & 3.96 & 4.05 & 3.05 & 4.03 & 4.11 \\
Pers. & 3.02 & 3.52 & 3.60 & 3.27 & 3.61 & 3.73 \\
\midrule
\tablegroup{7}{GLM-4.1V (9B)}
Info. & 3.63 & 3.88 & 4.17 & 3.75 & 4.07 & 4.38 \\
Read. & 4.54 & 4.90 & 4.96 & 4.62 & 4.92 & 4.95 \\
Sound. & 3.02 & 3.91 & 4.19 & 3.21 & 3.99 & 4.33 \\
Pers. & 3.12 & 3.63 & 4.09 & 3.36 & 3.74 & 4.17 \\
\bottomrule
\end{tabularx}
\end{table}

As shown in Table~\ref{tab:explainability}, automatic and human evaluations consistently rank ScamCourt highest across all four dimensions on both backbones.
Qwen's zero-shot explanations receive 4.86 for readability and 2.90 for soundness, illustrating the gap between fluent expression and evidence-supported reasoning.
SFT improves soundness substantially, and ScamCourt further raises the automatic scores from 3.96 to 4.05 on Qwen and from 3.91 to 4.19 on GLM, with human scores increasing from 4.03 to 4.11 and from 3.99 to 4.33, respectively.
The gains over SFT are more pronounced on GLM, particularly in persuasiveness, while readability remains similar between the fine-tuned settings.
The two evaluation modes support the observed gains in evidential grounding and explanatory relevance.

\section{Limitations}
\label{sec:limits}
Our study has several limitations. MechScam draws on existing video collections and may therefore underrepresent emerging scam mechanisms as online fraud continues to evolve. Subtle differences in communicative intent and policy interpretation can remain challenging to capture consistently. Moreover, our data and evaluation focus on YouTube; extending the benchmark to other platforms and incorporating richer temporal information are promising directions for future work.

\section{Ethical Considerations}
\label{sec:ethics}
This study analyzes publicly available YouTube videos without directly interacting with their creators or viewers. During data collection, we adhered to YouTube API rate limits to minimize additional load on the platform. Our annotations are content-level research judgments; legal culpability and financial outcomes require independent verification. The misleading arguments support robustness research and carry dual-use risks through their potential use in misleading claims or defenses. Any release of source identifiers and derived materials should respect applicable privacy, licensing, and redistribution requirements.

\end{document}